%% file: main.tex
\documentclass{article}

\input{_default_preamble}
\input{_custom_preamble}

\title{TopK Language Models}

\author{%
    Ryosuke Takahashi\thanks{Correspondence to \url{ryosuke.takahashi@dc.tohoku.ac.jp}.} \\
    Tohoku University / RIKEN\\
    \And 
    Tatsuro Inaba \\
    Tohoku University \\
    \And
    Kentaro Inui \\
    MBZUAI / Tohoku University / RIKEN \\
    \And
    Benjamin Heinzerling \\
    RIKEN / Tohoku University
}

\begin{document}

\maketitle

\begin{abstract}
  Sparse autoencoders (SAEs) have become an important tool for analyzing and interpreting the activation space of transformer-based language models (LMs).
  Trained to ``disentangle'' dense activation vectors represented in the neuron basis into sparse vectors represented in a learned overcomplete basis, SAEs can discover highly interpretable representations of concepts.
  However, SAEs suffer several shortcomings that diminish their utility and internal validity.
  Since SAEs are trained post-hoc, it is unclear if the failure to discover a particular concept is a failure on the SAE's side or due to the underlying LM not representing this concept.
  This problem is exacerbated by training conditions and architecture choices affecting which features an SAE learns.
  When tracing how LMs learn concepts during training, the lack of feature stability also makes it difficult to compare SAEs features across different checkpoints.
  To address these limitations, we introduce a modification to the transformer architecture that incorporates a TopK activation function at chosen layers, making the model's hidden states equivalent to the latent features of a TopK SAE. 
  This approach eliminates the need for post-hoc training while providing interpretability comparable to SAEs. 
  The resulting TopK LMs offer a favorable trade-off between model size, computational efficiency, and interpretability.
  Despite this simple architectural change, TopK LMs maintain their original capabilities while providing robust interpretability benefits. 
  Our experiments demonstrate that the sparse representations learned by TopK LMs enable successful steering through targeted neuron interventions and facilitate detailed analysis of neuron formation processes across checkpoints and layers. 
  These features make TopK LMs stable and reliable tools for understanding how language models learn and represent concepts, which we believe will significantly advance future research on model interpretability and controllability.
\end{abstract}

\section{Introduction}

\input{figure/overview}

Sparse autoencoders \citep[SAEs;][]{sharkey2023features,cunningham2023sparse,templeton2024scaling}, in particular TopK SAEs \citep{gao2024scaling}, have become an important tool for interpreting the internal representations of language models \citep{lieberum2024gemmascope,he2024llamascope,karvonen2025saebench,shu2025survey}.
However, they suffer from several shortcomings that diminish their utility and internal validity.
SAEs are trained post-hoc and are lossy in the sense that they do not perfectly reconstruct their input, leading to  a tradeoff between sparsity and reconstruction fidelity \citep{rajamanoharan2024gated,gao2024scaling}.
Post-hoc training makes it difficult to determine if the absence of a feature, say, a feature encoding the concept \emph{Golden Gate Bridge} \citep{templeton2024scaling}, is (a) due to the SAE failing to discover this feature during training, or (b) due to the lack of representation in the underlying LM, that is the LM may not have learned to represent \emph{Golden Gate Bridge} as a distinctly discoverable activation pattern \citep{leask2025canonical}.
This problem is exacerbated by the fact that the choice of a specific SAE architecture determines which kind of concept representations can be learned and which cannot, thereby restricting the set of discoverable features if the chosen SAE architecture does not match the LM's internal representations of the concepts of interest \citep{hindupur2025projecting}.
Yet another issue arising from the need for post-hoc training is high sensitivity to initial conditions, as SAEs trained with different random seeds but on the same underlying LM activations have been found to learn different features \citep{paulo2025sparse}.
All of the above shortcomings are compounded when using SAEs to analyze model training dynamics, e.g., requiring extra steps to encourage, but not guarantee, feature alignment across SAEs trained on different model checkpoints \citep{xu2025tracking} or limiting the analysis of cross-checkpoint SAE features to summary statistics \citep{inaba2025tracing}.

In this work, our goal is to devise an LM architecture that enjoys the interpretability of SAEs, but does not suffer the drawbacks of post-hoc training.
Specifically, we achieve SAE-like sparsity by simply replacing the final activation function  of selected layers in a transformer-based LM with the TopK activation function (see definition in \cref{subsec:topk_activation}), which has been shown to be a simple but highly effective sparsity constraint \citep{gao2024scaling}.
Having explored varying degrees of sparsity, placement of TopK-sparse layers, and model size, we found that it is possible to train a well-performing LM in which all but the last two layers are TopK sparse (\cref{fig:overview}, left).
To compensate for high sparsity, we increase layer width, i.e., hidden state dimensionality, but retain the width of feed-forward layers to keep the increase in the number of model parameters small.
In addition, we find that annealing the TopK sparsity over the course of training greatly improves convergence.
The resulting sparsely-activated LMs perform comparably to densely activated ones while offering SAE-style interpretability, which we demonstrate in three interpretability showcase studies (\cref{fig:overview}, right), namely (1) identifying neurons that are strongly associated with concepts such as ``work'', (2) single-neuron concept steerability, e.g., amplifying the ``work neuron'' steers the model towards generating texts about work, and (3) tracing neuron specialization during training and across layers, e.g., we can trace precisely how``work neuron'' forms across checkpoints, as well as discovering that, once formed, this neuron corresponds to a single dimension in the model's residual stream across all layers.

In summary, this work makes the following contributions:

\begin{itemize}
    \item We propose TopK LM, a sparsely activated transformer-based LM architecture. In this architecture, sparse activations are achieved by replacing the activation function in chosen layers with the TopK function.
    \item Systematically training various instantiations of our proposed architecture, we study the impact of factors such as model size, degree of TopK sparsity and number of TopK layers on language modeling and downstream task performance. Our results demonstrate a favorable trade-off between sparsity and model size, i.e., sparsely-activated TopK LMs perform well with high degrees of sparsity with only a moderate increase in model size compared to otherwise identical dense models.
    \item We identify sparsity annealing as a simple and efficient method for improving TopK model training.
    \item We showcase multiple examples of SAE-like interpretability of TopK LMs, namely highly interpretable features, concept steerability, and high traceability of neuron specialization during training.
    \item In a broader context, we see TopK LMs as an exciting example of inherently interpretable neural networks and argue that recent advances such as TopK-SAEs are a great opportunity to revisit the tradeoff between interpretability and performance.
\end{itemize}

\section{TopK Language Model}

\subsection{Architecture}
\label{sec:topk_lm_architecture}

In this section we describe the TopK LM architecture, which introduces activation sparsity by inserting TopK activation functions in chosen layers, as illustrated in comparison to a baseline LM in \cref{fig:topk-lm}.

\paragraph{TopK activation}
\label{subsec:topk_activation}
Let $x = (x_1, \dots, x_d)\in\mathbb{R}^d$. 
and denote by $\tau_{k}(x)$ the $k$-th largest value in $\{x_1,\dots,x_d\}$,
where $k$ is a hyperparameter that controls the sparsity level by selecting how many activations are retained. 
The TopK activation function $\mathcal{T}_k\colon\mathbb{R}^d\to\mathbb{R}^d$ is defined component‐wise as
\begin{align}
    y_i = 
    \begin{cases}
        f(x_i) & x_i \ge \tau_{k}(x)\\
        0      & \text{otherwise}
    \end{cases}
\end{align}
or equivalently
\begin{align}
    y_i = f(x_i)\,\mathbf{1}_{\{\,x_i \ge \tau_{k}(x)\}}\quad i = 1,\dots,d
\end{align}
where $f$ is an element-wise nonlinearity (e.g., ReLU).

\input{figure/topk_lm}

\paragraph{Annealed TopK smoothing}
Aiming to improve training stability and convergence, we experimented with varying the degree of sparsity throughout training.
Based on the fact that current LM training methods and received best practices are optimized for dense models, we hypothesized that it might be beneficial to gradually transition from a fully dense to a TopK-sparse model.
Empirically, we found that an annealed TopK activation works well, thereby confirming this hypothesis.
Specifically, the annealed TopK function gradually transitions from dense to sparse activations, via an annealing factor $\alpha\in[0,1]$ that decays from 1 to 0 over the course of training.
In preliminary experiments we found linear decay to work well and delegate exploring more sophisticated decay schedules to future work.
The linearly annealed TopK activation is then
\begin{align}
    y = \alpha\,f(x) + (1-\alpha)\bigl(f(x)\odot\mathbf{1}_{\{\,x \ge \tau_k(x)\}}\bigr)
\end{align}
where $f(x) = (f(x_1),\dots,f(x_d))$ and $\odot$ denotes element-wise product.
In words, at the start of training ($\alpha=1$), the block is fully dense, for $\alpha\to0$ the non-TopK activation becomes smaller, and with $\alpha=0$ this activation function is identical to TopK.
This approach is related to, but distinct from, dense-to-sparse training \citep{Kusupati-2020-soft-threshold,Graesser-2022-sparse-training}.

\paragraph{Hybrid block placement}
To preserve representational power in the higher layers while still benefiting from sparsity, we introduce a hyperparameter $n_{\text{nontopk}}$ to denote the number of final Transformer layers that remain in their standard (dense) form. 
Concretely, given a total of $L$ layers, we replace the first $ L-n_{\text{nontopk}}$ layers with TopK blocks and leave the last $n_{\text{nontopk}}$ layers as conventional Transformer blocks. 
This hybrid design allows us to exploit sparsity for most of the network's depth while retaining full expressivity in the final $n_{\text{nontopk}}$ layers.

\subsection{Pre-training Settings}
\label{sec:pre_training_settings}

In this section we describe the training settings for the TopK LM introduced in \cref{sec:topk_lm_architecture}.

\paragraph{Model Configuration}
We conduct pre-training on two decoder-only variants of the Llama architecture~\citep{touvron-2023-llama}: a vanilla baseline and TopK-LM, which integrates the TopK mechanism. 
Both variants are instantiated with hidden dimensionality $D\in\{1024,2048\}$ and transformer depth $L\in\{8,16,24\}$, employing $D/128$ attention heads in every configuration. 
In the TopK-LM variant, each layer retains only the top $k=64$ activations, preserves the final $n_{\mathrm{lastnontopk}}=2$ positions from masking, and applies a masking strength that increases linearly during the first $20\%$ of training steps (annealing-step-ratio = $0.2$).

\paragraph{Tokenizer and Dataset}
All models share the official Llama tokenizer, which uses a fixed vocabulary of $32000$ word-piece tokens to ensure consistent input processing.
We use FineWeb Edu corpus~\citep{penedo-2024-fineweb, lozhkov-2024-fineweb-edu} as the pre-training data, and he amount of training data is approximately $20$ billion tokens.
A held-out portion of the same corpus serves as the validation set.

\paragraph{Training Hyperparameters}
Optimization is performed with AdamW ~\citep{loshchilov-2018-decoupled-weight-decay} using an initial learning rate of $3\times 10^{-4}$, $(\beta_{1},\beta_{2})=(0.9,0.95)$, $\epsilon=10^{-8}$, and weight decay of $0.1$. 
We employ a linear warmup over the first $10\%$ of total training steps, followed by cosine decay of the learning rate. 
Gradient norms are clipped to a maximum of $10.0$ to ensure numerical stability. 
Training uses a global batch size of $1024$, with $32$ sequences per GPU across eight NVIDIA H200 SXM cards and gradient accumulation over four micro-batches, yielding an effective large-batch regime. 
Input sequences are truncated or padded to a maximum of $1024$ tokens. 
The total number of training steps is $20000$, corresponding to the processing of approximately $2.0\times 10^{10}$ tokens.

\paragraph{Implementation and Hardware Environment}
Our implementation builds upon Meta's ``Meta Lingua'' PyTorch LLM library~\citep{videau-2024-meta-lingua}.
All experiments are conducted on a single node with eight NVIDIA H200 SXM GPUs, each with $ 141$ GB of VRAM.

\paragraph{Distributed Training and Optimization Settings}
We employ PyTorch's Fully Sharded Data Parallel~\citep{zhao-2023-pytorch-fsdp} to fully shard model parameters, gradients, and optimizer states across the eight GPUs. 
No tensor parallelism is used. 
Forward and backward computations are executed in mixed precision using the bfloat16 format.

\subsection{Evaluation}
Following previous research~\cite{yano-2025-step}, We evaluate each pre-trained model along two axes---linguistic fluency via perplexity and zero-shot generalization via accuracy---using the lm-evaluation-harness~\citep{gao-2024-eval-harness}.

First, we report perplexity on the FineWeb-Edu validation split and WikiText~\citep{merity-2017-pointer}. 
In addition, for LAMBADA~\citep{paperno-2016-lambada}, we compute both perplexity and accuracy to capture performance on a narrative cloze task requiring long context.

Second, we measure zero-shot accuracy on a suite of multiple-choice benchmarks spanning commonsense reasoning and question answering, including WinoGrande~\citep{sakaguchi-2021-winogrande}, HellaSwag~\citep{zellers-2019-hellaswag}, both the Easy and Challenge sets of ARC~\citep{clark-2018-think}, and OpenBookQA~\citep{mihaylov-2018-suit}.
For all choice-based tasks, we report accuracy.  
These evaluations provide a comprehensive portrait of each model's generative fluency and emergent reasoning capabilities.

\cref{tab:eval_results} compares the baseline and our TopK-LM across model sizes with hidden dimension $D\in{1024,2048}$ and layer counts $L\in{8,16,24}$\footnote{During the evaluation, the context length is set to 256.}. 
Although the TopK mechanism does raise perplexity modestly, for instance, at $D=1024, L=24$ validation perplexity increases from $11.76$ to $14.96$ and LAMBADA perplexity from $49.29$ to $53.16$ while end-task accuracies remain largely on par with the baseline. 
In that same configuration, the TopK LM surpasses the baseline on LAMBADA accuracy ($33.32\%$ vs.\ $31.61\%$) and Winogrande accuracy ($49.57\%$ vs.\ $48.93\%$), and incurs only small drops on ARC and OBQA. 
For the larger $D=2048, L=24$ model, TopK LM's validation perplexity of $11.63$ (vs. $9.33$) is accompanied by competitive accuracies across all six benchmarks, with differences of just a few points. 
These results confirm that enforcing a TopK sparse‐activation constraint does not materially degrade language‐model performance.

\input{table/eval_results}

\section{Interpretability Showcase: Neuron Specialization in a Pretrained TopK LM}

\subsection{Token Entropy}
To measure how ``spread out'' a neuron's activations are over the vocabulary in a given hidden layer, we employ Token Entropy.  
Let $N$ be the total number of tokens in the corpus, $|V|$ the vocabulary size, $\ell$ the layer index, and $k$ the neuron (hidden‐dimension) index.  
Denote by $d_i\in{1,\dots,|V|}$ the token ID at position $i$ in the corpus, and by $h_{\ell,k}(i)$ the activation of neuron $k$ at layer $\ell$ for that token.  
We first compute the average activation of neuron $k$ in layer $\ell$ to token $d$ as
\begin{align}
\label{eq:token_entropy_mu}
    \mu_{\ell,k,d} &= \frac{1}{N}\sum_{i : d_i = d} h_{\ell,k}(i)
    \quad(d=1,\dots,\lvert V\rvert)
\end{align}
\cref{eq:token_entropy_mu} thus represents the mean response of neuron~$k$ to token~$d$.  We then normalize over all tokens to obtain a probability distribution $p_{\ell,k,\cdot}$:
\begin{align}
\label{eq:token_entropy_p}
    p_{\ell,k,d} &= \frac{\mu_{\ell,k,d}}       {\sum_{d'=1}^{\lvert V\rvert}\mu_{\ell,k,d'}}
    \quad(d=1,\dots,\lvert V\rvert)
\end{align}
Finally, the token entropy of neuron $k$ in layer $\ell$ is defined by
\begin{align}
\label{eq:token_entropy}
    H_{\mathrm{token}}(\ell,k) &= -\sum_{d=1}^{\lvert V\rvert}p_{\ell,k,d}\,\log p_{\ell,k,d}
\end{align}
A small value of $H_{\mathrm{token}}(\ell,k)$ indicates that neuron $k$ is highly token‐selective, exhibiting strong activation only for a narrow subset of the vocabulary.  
Conversely, a large value (up to $\log|V|$) implies that its activation is distributed more uniformly across the entire vocabulary, i.e., the neuron is more generic.

\subsection{Semantic Entropy}
\label{sec:semantic_entropy}
In parallel with Token Entropy, we quantify how spread out a neuron's activations are over semantically coherent clusters of the vocabulary by defining \emph{Semantic Entropy}.  
While prior work has proposed sentence-level measures of semantic uncertainty \citep{kuhn-2023-semantic-uncertainty,farquhar-2024-detecting-hallucinations}, we adapt this idea to the neuron level, thereby enabling a quantitative assessment of which meaning-related token sets each hidden neuron selectively responds to.
Let $\ell$ denote the layer index and $k$ the neuron index.  We begin by identifying, for each neuron $(\ell,k)$, the set of tokens whose mean activation exceeds a threshold.
Denote by $A^{(\ell)}{k,d}$ the average activation of neuron $(\ell,k)$ to token $d$, and let $\theta$ be 70\% of the $99.9$th-percentile of all $A^{(\ell)}_{k,d}$ values.  
We then define the selected vocabulary subset:
\begin{align}
\label{eq:semantic_entropy_v}
V_{\ell,k} = \{\,d \mid A^{(\ell)}_{k,d} > \theta\}
\end{align}
Next, to capture semantic coherence among the tokens in $V_{\ell,k}$, we retrieve each token's embedding vector $e_d$ and compute pairwise cosine similarities:
\begin{align}
\label{eq:semantic_entropy_cos}
S_{d,d'} = \cos(e_d, e_{d'}) \quad \forall d,d' \in V_{\ell,k}
\end{align}
Since more frequent tokens should contribute proportionally more to the semantic structure, we weight each token pair $(d,d')$ by the product of their corpus frequencies $f_d$ and $f_{d'}$:
\begin{align}
\label{eq:semantic_entropy_w}
w_{d,d'} = f_d \cdot f_{d'} \quad \forall d,d' \in V_{\ell,k}
\end{align}
We then partition the range of cosine similarities into $n_{\mathrm{bins}}$\footnote{In our experiments, we set $n_{\mathrm{bins}}=1000$.} contiguous bins ${B_i}$ and form a weighted histogram distribution over these bins:
\begin{align}
\label{eq:semantic_entropy_p}
p_i = \sum_{(d,d') \in B_i} \frac{w_{d,d'}}{\sum_{d,d' \in V_{\ell,k}} w_{d,d'}}
\end{align}
Finally, the Semantic Entropy of neuron $(\ell,k)$ is defined as the Shannon entropy of this bin‐weighted similarity distribution:
\begin{align}
\label{eq:semantic_entropy}
H_{\mathrm{sem}}(\ell,k) = -\sum_{i=1}^{n_{\text{bins}}} p_i \log_2 p_i
\end{align}
A small value of $H_{\mathrm{sem}}(\ell,k)$ indicates that the neuron's high activations are concentrated on a semantically coherent subset of tokens, and hence that the neuron is \emph{semantic-selective}.  
Conversely, a large value (up to $\log_2 n_{\mathrm{bins}}$) implies that its activations span multiple semantic clusters, indicating that the neuron exhibits a more generic response.

\subsection{Results: Entropy Comparison Between TopK and Baseline Models}
\label{sec:results_entropy}
\input{figure/layer24_entropy_comparison}

We analyze the impact of TopK activation ($k=64$, with the final two layers operating in a non TopK regime) on 24‐layer transformer models having hidden‐state dimensions of 1024 and 2048.  
\cref{fig:layer24_entropy_comparison} depicts the layer‐wise distributions of both token entropy and semantic entropy.

The standard (baseline) model (blue curve) exhibits a distinctive entropy profile.  
In the earliest layers (0--5), token entropy remains high (approximately 9.3--9.5), then drops sharply in layers 5--7, before gradually rising again to around 8--9 near the output layers.  
This pattern suggests a processing trajectory of broad feature capture in shallow layers, specialization in mid‐layers, and re‐abstraction in deeper layers.  
By contrast, the TopK activated models (green curves) maintain uniformly low entropy (6--7) across all layers, indicating stronger functional specialization of individual neurons.  Notably, the $d=2048$ variant (dark green) achieves even lower entropy than the $d=1024$ model (light green), implying that under a fixed TopK constraint a wider hidden dimension exploits the representation space more sparsely. 
The pronounced increase in entropy in the final two layers of the TopK models reflects the removal of the sparsification constraint in the non TopK regime.  
The shaded regions in the plot denote one standard deviation: the baseline model shows particularly large variance in mid‐layers (8--15), signaling substantial functional heterogeneity among neurons, whereas the TopK models exhibit narrower variance in token entropy, consistent with a homogenizing effect on neuron behavior.

A clear divergence between baseline and TopK models is also evident in semantic entropy. 
The standard model maintains relatively high values (4.3--4.4) with little layer‐to‐layer variation.  
In contrast, the TopK variants show markedly lower semantic entropy (3.0--3.7), with the $d=2048$ model again registering the lowest values.  As with token entropy, a sharp rise appears in semantic entropy in the last two (non TopK) layers, reflecting the reintegration and abstraction of information.  
The standard deviations of semantic entropy are substantial for both model types; in particular, the lower tails of the TopK models' variance indicate that some neurons are highly specialized---responding almost exclusively to a narrow, semantically coherent set of tokens---and thus function as semantic feature detectors.

\section{Interpretability Showcase: Concept Steering}

\input{table/steering_example}

In the previous section we saw correlational evidence suggesting that individual neurons in the TopK LM sparsely represent concepts in a manner similar to monosemantic features in SAEs. 
To establish a causal link between these single neurons and model outputs, we now perform targeted intervention experiments, following prior work on causal interventions on LMs~\citep{geiger-2021-causal,tamkin-2023-codebook}.
Specifically, we perform activation patching with the goal of concept steering \citep{conmy2024activation,templeton2024scaling,kim2025concept}.
First, we select neurons whose manipulation is likely to yield interpretable effects.  
Guided by the semantic entropy results from \cref{sec:semantic_entropy}, we focus on neurons exhibiting low semantic entropy, since these neurons respond selectively to specific concepts or attributes and are thus good candidates for concept steering.  
For each target neuron $(\ell,n)$, we examine the tokens that elicit the highest mean activation and manually assign an interpretable label (e.g., ``work,'' ``numbers,'' or ``history'') based on these token patterns. We leave automatic assignment of labels 
\citep{paulo2024automatically} to future work.
Our intervention consists of adding a constant offset $\delta$ to the activation of a single neuron at every sequence position.  
Formally, if $h_{\ell,n}(i)$ denotes the original pre-activation of neuron $n$ in layer $\ell$ at token position $i$, we replace it with $h'_{\ell,n}(i) = h_{\ell,n}(i)+\delta.$
Empirically, we find that offsets in the range $\delta\in[5,30]$ produce the most pronounced steering effects. 
For generation, we sample with temperature = 0.7, top-$p$ = 0.9, and top-$k$ = 50, generating up to 128 tokens.  
To observe the effect concept steering, we prompt the model with story-like prompt (``Once upon a time,''), following the experimental setup of \citep{tamkin-2023-codebook}.   
As illustrated in \cref{tab:steering_example}, steering a neuron associated with the concept ``work'' causes the model to produce markedly more work-related text, thereby demonstrating the causal influence of that neuron on generation.

\section{Interpretability Showcase: Tracing Neuron Specialization Across Training Checkpoints and Model Layers}

To trace how individual neurons acquire concept‐specific selectivity, we compare their activation statistics both across training checkpoints and along the depth of the network.  
\cref{fig:token_and_semantic_entropy_comparison_across_ckpt} shows the evolution of semantic entropy (and token entropy) during training for a 24‐layer TopK transformer LM.  
The horizontal axis indexes successive training checkpoints (in number of training steps), while the vertical axis reports average entropy values in hidden states. 
The light‐green curve corresponds to the last TopK‐active layer, and the dark‐green curve to the final non TopK layer.  We observe that the lower range of both token entropy and semantic entropy in the TopK layers decreases as training progresses, indicating that neurons develop increasingly specialized representations over time.

\input{figure/token_and_semantic_entropy_comparison_across_ckpt}

We further investigate two central questions about this formation process: (1) at which stage of training do specialized neurons first emerge, and (2) whether neurons with consistent semantic selectivity appear concurrently across multiple layers. 
Unlike SAE methods, the TopK‐based framework allows us to continuously track the same hidden‐dimension index both across checkpoints and across layers.

\cref{fig:layer24_neuron_activation_map_for_token_work} presents a concrete example of this dynamic.  
Using our concept‐steering procedure, we identify a neuron at layer 22, hidden dimension 894, that becomes specialized for the concept ``work.''  
The figure visualizes the activation strength of this dimension in response to the token ``work'' over successive checkpoints and layers.  
In the early training phase (up to $10^3$ steps), the dimension’s selective response to ``work'' is confined to lower layers.  
As training continues, the same dimension exhibits strong, selective activation for ``work'' across nearly all layers. 
This behavior confirms that our TopK analysis method can effectively track the emergence and deepening of concept‐specific neuron selectivity both in depth and over the course of training, showing an example of progressive specialization in a language model.

\input{figure/layer24_neuron_activation_map_for_token_work}

\section{Related Work}
In the context of neural networks, \emph{sparsity} has several related but distinct meanings \citep{farina2024sparsity}: sparse model weights \citep{louizos2018learning,mocanu2018scalabe,evci2020rigging,jayakumar2021kast,hunter2021two}, sparse attention \citep{child2019long,gupta2021memory}, sparse activation of components in Mixture-of-Experts-style architectures \citep{shazeer2017outrageously,jaszczur2021sparse}, and SAE-style representational sparsity of hidden state activations \citep{makhzani2015winner,ahmad2019dense,hunter2021two,sharkey2023features,cunningham2023sparse,templeton2024scaling,gao2024scaling}. Our work falls into the last category. Closest among related work on representational sparsity are codebook feature layers \citep{tamkin-2023-codebook}, which, similar to our TopK layers, also enjoy inherent interpretability, but differ in the way sparsity is achieved, since their method relies on vector quantization and maximum inner product search, while our proposed architecture requires only minimal modification of the transformer architecture.

\section{Conclusion}
We have introduced TopK LMs, a novel architecture that directly incorporates sparsity into transformer LMs through a TopK activation mechanism. Unlike post-hoc sparse autoencoders, our approach makes interpretability an intrinsic property of the model while addressing key limitations of SAEs. Our experiments demonstrate that TopK LMs achieve competitive performance with only moderate increases in model size.
TopK LMs offer substantial interpretability benefits: neurons exhibit higher concept specialization, enable effective concept steering, and allow continuous tracing of neuron development across both training checkpoints and model layers. This provides unprecedented insights into how LMs learn to represent concepts during training.
By making interpretability a built-in feature rather than a post-hoc analysis tool, TopK LMs represent a significant advance toward trustworthy AI systems. They demonstrate that performance and interpretability need not be competing objectives, opening promising directions for future research on transparent, controllable language models.

\section*{Acknowledgments}
This work was supported by JST CREST, Japan Grant Number JPMJCR20D2 and AMED, Grant Number JP25wm0625405 and JSPS KAKENHI, Grant Number JP25KJ06300 and JST BOOST, Grant Number JPMJBS2421.

\bibliography{custom}
\bibliographystyle{plain}

\clearpage
\appendix

\section{Ablation Study}
In this section, we investigate the impact of key TopK LM hyperparameters on model performance and the sparsity–accuracy trade‐off. Specifically, we conduct three sets of experiments:
\begin{itemize}
\item Adjusting the number of non‐TopK (fully dense) layers to assess how the balance between sparse and dense components influences overall results (\cref{tab:ablation_nontopk}).
\item Varying the TopK–$k$ value to measure its effect on perplexity and activation sparsity (see \cref{tab:ablation_topk_k}).
\item Changing the hidden‐state dimensionality to evaluate scalability and convergence behavior across different model capacities (\cref{tab:ablation_hidden_dim}).
\end{itemize}
All experiments follow the training protocol described in \cref{sec:pre_training_settings}, ensuring fair comparison. 
The results highlight which hyperparameter settings offer the best compromise between interpretability, computational cost, and downstream performance.

\input{table/ablation_nontopk}

\input{table/ablation_topk_k}

\input{table/ablation_hidden_dim}

\section{Additional Results}
In Section \ref{sec:results_entropy}, we compared token‐level and semantic entropies between the baseline and TopK models on our 24‐layer configuration. 
To verify that these findings generalize across different depths, we show the corresponding entropy comparisons for 8‐layer and 16‐layer models in \cref{fig:layer8_entropy_comparison,fig:layer16_entropy_comparison}, respectively. 
In both the 8‐layer and 16‐layer cases, we observe the same pattern as in the 24‐layer experiments: TopK activation consistently reduces token and semantic entropy relative to the dense baseline, indicating stronger neuron specialization.

\input{figure/layer8_entropy_comparison}

\input{figure/layer16_entropy_comparison}

\section{Limitations}
\label{sec:limitations}
While our TopK LM architecture shows promising results for building interpretable foundation models, several important limitations should be addressed in future work.

\paragraph{Scale Limitations}
Our experiments were conducted primarily on 24-layer models with 2048-dimensional hidden states ($\approx$ 1B parameters). The effectiveness of TopK sparsity on significantly larger models (tens or hundreds of billions of parameters) remains an open question. As model scale increases, the optimal sparsity levels, TopK-k values, and placement of sparse layers may need to be recalibrated. Additionally, the computational overhead of TopK operations might become more pronounced at extreme scales, potentially requiring algorithmic optimizations.

\paragraph{Limited Evaluation Domains}
Our evaluation focused primarily on language modeling perplexity and standard reasoning benchmarks. This leaves important questions about how TopK LMs perform on more specialized downstream tasks, particularly those requiring fine-grained knowledge integration, complex reasoning, or domain-specific expertise. Further evaluation across diverse task categories would provide a more complete picture of the tradeoffs between sparsity and task performance.

\paragraph{Long-Context Performance}
We have not thoroughly examined how TopK sparsity affects long-context modeling capabilities. As context length becomes increasingly important for modern LLM applications, understanding how sparse activations impact information retention across thousands or tens of thousands of tokens is critical. The interaction between TopK sparsity and attention mechanisms over long sequences deserves particular investigation.

\paragraph{Multimodal Extensions}
Our approach is currently limited to text-only models. Extending TopK activation to multimodal architectures---combining text with images, audio, or other modalities---represents an important direction for future research. Multimodal models may require different sparsity patterns across modalities or novel architectural adaptations to maintain performance while preserving interpretability.

\paragraph{Societal Applications}
While we present a framework for more interpretable foundation models, we have not explored specific societal applications that could benefit from this interpretability. Future work should investigate how TopK LMs can address concrete challenges in areas such as model alignment, fairness assessment, safety evaluation, and trustworthy AI deployment. Converting the technical advantages of interpretability into practical societal benefits remains an important next step.
Addressing these limitations will be crucial for establishing TopK LMs as a broadly applicable approach to interpretable foundation models and realizing their potential across diverse applications and deployment scenarios.

\section{Impact Statement}
\label{sec:impact_statement}

As we described in \cref{sec:limitations}, we did not extensively explore the societal applications of these models.
However, the inherent interpretability of TopK LMs has the potential to ensure that their decision-making processes are transparent.
It must promote ethical use and mitigate risks.

In contrast, a potential negative impact in the future could arise if these interpretable methods are exploited to extract sensitive data, such as personally identifiable information (PII), from pre-trained language models.
To the best of our knowledge, no such misuse is currently possible, and we believe that this research does not pose any direct negative impact at present.

\end{document}

%% file: _default_preamble.tex
\PassOptionsToPackage{numbers, compress}{natbib}

\usepackage[preprint]{neurips_2025}

\usepackage[utf8]{inputenc} %
\usepackage[T1]{fontenc}    %
\usepackage{hyperref}       %
\usepackage{url}            %
\usepackage{booktabs}       %
\usepackage{amsfonts}       %
\usepackage{nicefrac}       %
\usepackage{microtype}      %
\usepackage{xcolor}         %

%% file: _custom_preamble.tex
\def\VersionWithComments{}
\usepackage{adjustbox}
\ifdefined\VersionWithComments%
    \usepackage[colorinlistoftodos]{todonotes}
\else
    \usepackage[disable]{todonotes}
\fi

\setuptodonotes{inline}

\newlength{\needcitelength}
\newcommand{\needcite}[1]{\settowidth{\needcitelength}{cites: {#1}}\todo[noinlinepar,inlinewidth=\the\needcitelength,size=\scriptsize]{cites: {#1}}}
\newcommand{\needref}[1]{\settowidth{\needcitelength}{ref: {#1}}\todo[color=yellow,noinlinepar,inlinewidth=\the\needcitelength,size=\scriptsize]{ref: {#1}}}

\usepackage[whole]{bxcjkjatype}

\usepackage{amsmath}
\usepackage{amssymb}
\usepackage{amsfonts}
\usepackage{bm}

\usepackage[capitalise,english,nameinlink]{cleveref} %
\Crefname{figure}{\text{Figure}}{\text{Figures}}
\creflabelformat{figure}{#2\textup{#1}#3}
\Crefname{equation}{\text{Equation}}{\text{Equations}}
\creflabelformat{equation}{#2\textup{#1}#3}

\usepackage{mathrsfs}
\usepackage{mathtools}
\usepackage{multirow}
\usepackage{dcolumn}
\usepackage{comment}
\usepackage{wrapfig}

\usepackage{enumitem}
\setlist{noitemsep, topsep=2pt}
\setenumerate{itemsep=3pt,topsep=2pt}

\definecolor{darkblue}{rgb}{0, 0, 0.5}
\hypersetup{colorlinks=true, citecolor=darkblue, linkcolor=darkblue, urlcolor=darkblue}

\usepackage[normalem]{ulem}
\useunder{\uline}{\ul}{}
\usepackage{pifont}
\newcommand{\mycheckmark}{\ding{51}}
\newcommand{\myxmark}{\ding{55}}

%% file: figure/overview.tex
\begin{figure}[t]
\centering
\includegraphics[width=\linewidth]{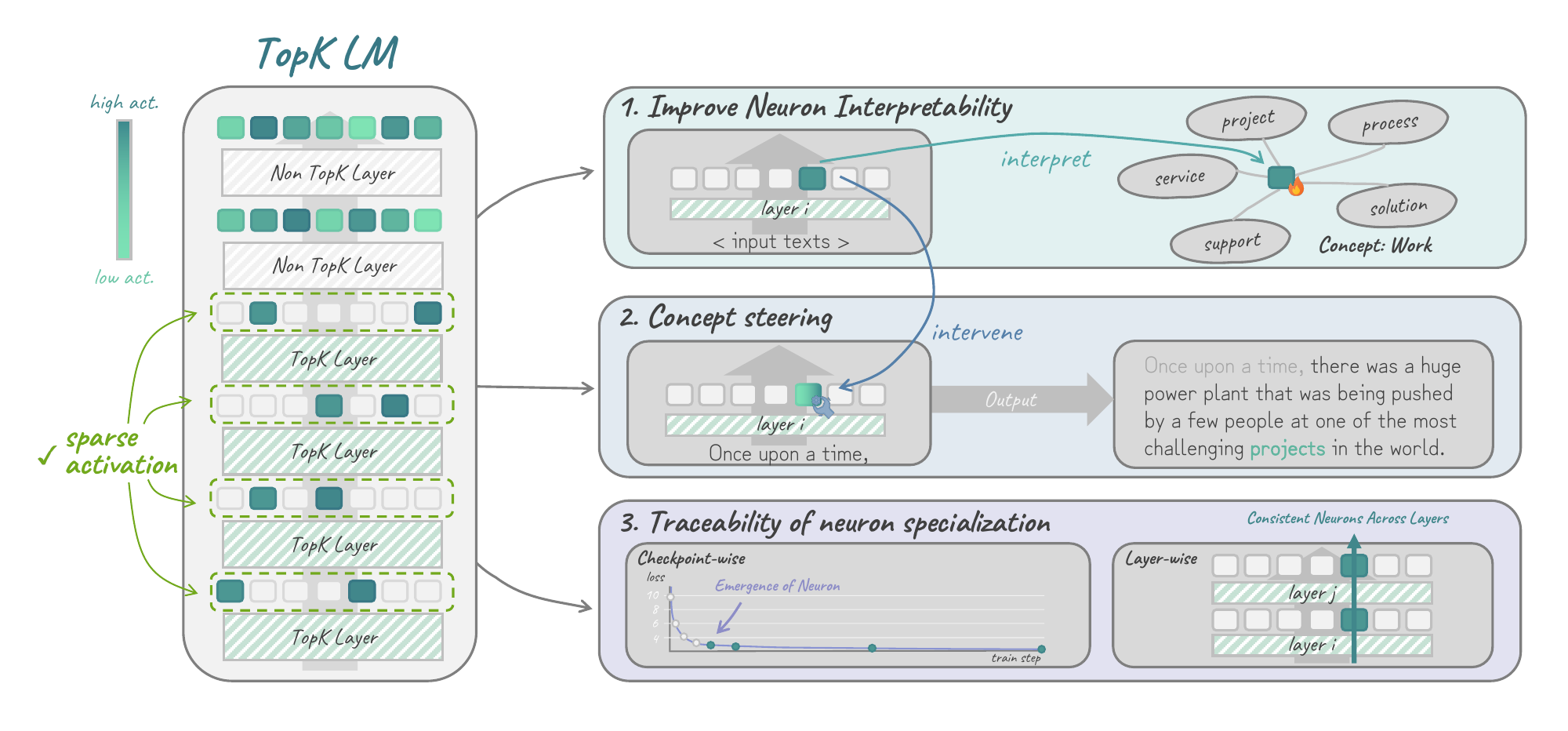}
\caption{
\textbf{Overview of the TopK LM architecture and its interpretability advantages.}
Left: The architecture integrates TopK activation functions in selected layers to achieve sparse activation patterns.
Right: TopK LMs enjoy SAE-like interpretability. (1) Improved neuron interpretability: individual neurons can be clearly interpreted as representing specific concepts (e.g., "Work"). (2) Steerability: single-neuron interventions can steer text output towards specific concepts, e.g. ``work''. (3) Traceability: The formation of specialized neurons can be traced across training checkpoints and across different layers.
That is, TopK LM combines the performance benefits of transformer-based LMs with the interpretability advantages of sparse autoencoders, without requiring post-hoc training.
}
\label{fig:overview}
\end{figure}

%% file: figure/topk_lm.tex
\begin{wrapfigure}{r}{0.5\textwidth}
  \centering
  \includegraphics[width=0.45\textwidth]{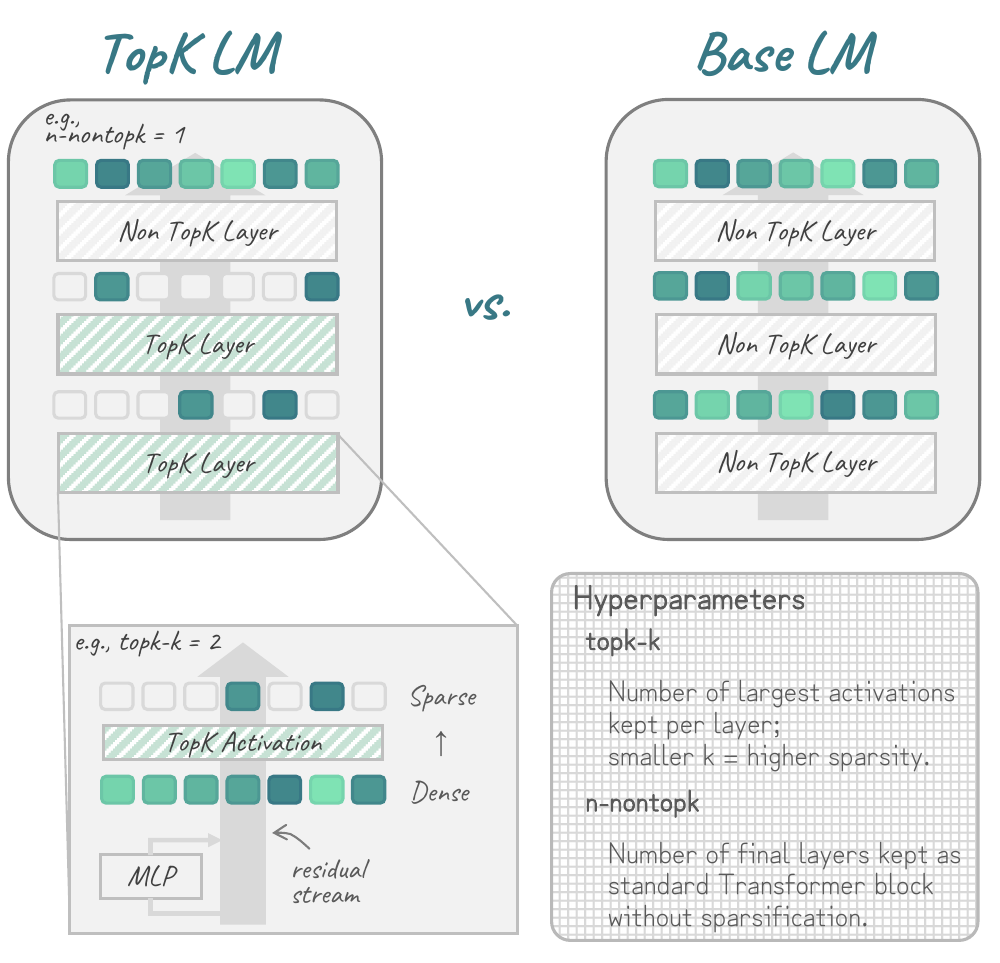}
  \caption{TopK LM: A sparse neural network architecture that selectively activates only the k-largest values in early layers while maintaining expressivity through dense processing in final layers.}
  \label{fig:topk-lm}
\end{wrapfigure}

%% file: table/eval_results.tex
\begin{table}[t]
\centering
\small
\caption{
Perplexity (\textdownarrow) on validation, LAMBADA and Wikitext datasets, alongside zero‐shot accuracy (\textuparrow) on six downstream benchmarks (LAMBADA, ARC‐Easy, ARC‐Challenge, Winogrande, OpenBookQA, HellaSwag), for baseline (Non TopK) and TopK LM models with hidden dimensionality $D\in\{1024,2048\}$ and layer depths $L\in\{8,16,24\}$.
}
\begin{adjustbox}{max width=\linewidth}
    \begin{tabular}{lc|ccc|cccccc}
        \toprule
        \multirow{2}{*}{Layers} & \multirow{2}{*}{TopK} & \multicolumn{3}{c|}{Perplexity \textdownarrow} & \multicolumn{6}{c}{Accuracy \textuparrow} \\ 
        \cmidrule(lr){3-5} \cmidrule(lr){6-11} 
         &  & Valid. & LAMBADA & Wikitext & LAMBADA & ARC-e & ARC-c & Winogrande & OBQA & HellaSwag \\ 
        \midrule
        \multicolumn{11}{l}{\hspace{-0.22cm} {\ul hidden dim: 1024}} \\[0.5em]
        8 & \myxmark & 14.67 & 112.80 & 20.96 & 23.71 & 50.00 & 20.39 & 50.67 & 18.20 & 28.77 \\
        16 & \myxmark & 12.67 & 67.37 & 17.69 & 27.87 & 55.56 & 22.18 & 49.88 & 19.80 & 30.86 \\
        24 & \myxmark & 11.76 & 49.29 & 15.99 & 31.61 & 56.27 & 24.06 & 48.93 & 21.80 & 32.11 \\
        8 & \mycheckmark & 16.13 & 74.60 & 71.23 & 30.16 & 30.93 & 20.05 & 50.12 & 26.80 & 26.60 \\
        16 & \mycheckmark & 15.25 & 65.36 & 42.72 & 30.99 & 29.76 & 21.84 & 49.41 & 26.40 & 25.92 \\
        24 & \mycheckmark & 14.96 & 53.16 & 25.93 & 33.32 & 30.51 & 22.61 & 49.57 & 23.60 & 25.95 \\
        \midrule
        \multicolumn{11}{l}{\hspace{-0.22cm} {\ul hidden dim: 2048}} \\[0.5em]
        8 & \myxmark & 11.22 & 43.69 & 15.04 & 31.88 & 58.08 & 23.98 & 51.14 & 22.40 & 32.76 \\
        16 & \myxmark & 9.89 & 26.90 & 12.91 & 37.24 & 61.32 & 27.22 & 51.70 & 22.80 & 35.71 \\
        24 & \myxmark & 9.33 & 22.24 & 11.73 & 38.81 & 64.06 & 29.52 & 53.20 & 25.40 & 37.49 \\
        8 & \mycheckmark & 12.32 & 46.61 & 46.94 & 27.71 & 33.71 & 22.01 & 49.72 & 26.60 & 26.76 \\
        16 & \mycheckmark & 11.77 & 35.48 & 20.34 & 30.41 & 35.52 & 20.82 & 52.09 & 27.20 & 27.69 \\
        24 & \mycheckmark & 11.63 & 31.91 & 21.98 & 35.11 & 32.62 & 21.42 & 52.33 & 27.80 & 27.90 \\
        \bottomrule
    \end{tabular}
\end{adjustbox}
\label{tab:eval_results}
\end{table}

%% file: figure/layer24_entropy_comparison.tex
\begin{figure}[t]
\centering
\includegraphics[width=\linewidth]{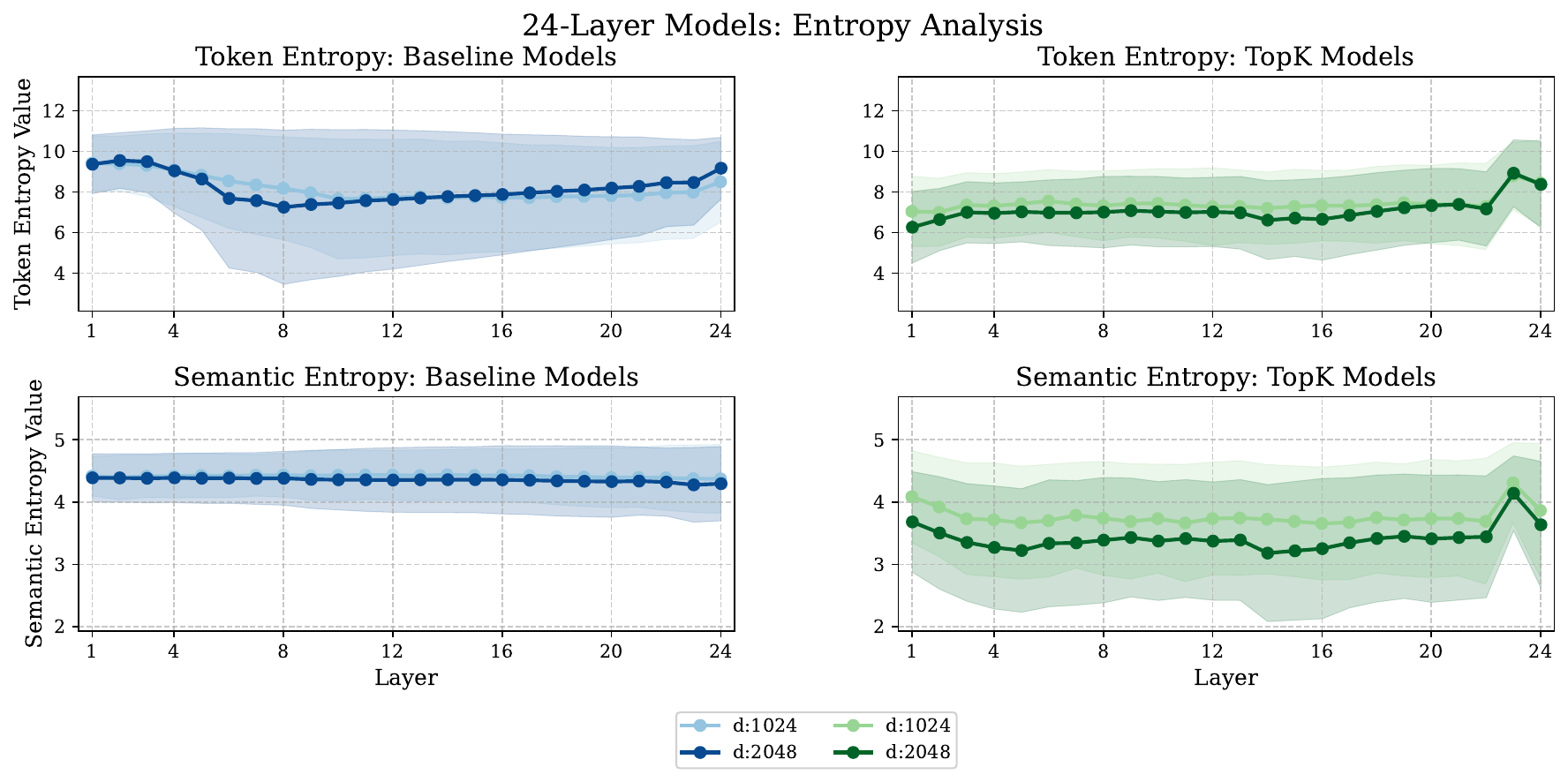}
\caption{
Baseline vs TopK models (24-layer). (Left) Token Entropy. (Right) Semantic Entropy. 
The horizontal axis indexes model layers; the solid curves plot the mean entropy at each layer, and the shaded bands denote one standard deviation around the mean.
}
\label{fig:layer24_entropy_comparison}
\end{figure}

%% file: table/steering_example.tex
\begin{table}[t]
\centering
\small
\caption{Effect of concept steering. By amplifying the activation of a single neuron, we can steer a TopK LM towards generating text about the concept associated with this neuron. The first row (Concept: \emph{none}) shows the model's baseline completion for the prompt ``Once upon a time'' without any amplification. Each of the remaining rows shows the effect of amplifying a specific neuron (identified as \emph{layer:index}). For example, amplifying neuron 21:1957, i.e., the 1957th entry of the activation vector of layer 21,  steers the model towards generating text about energy.}

\begin{tabular}[h!]{llp{9.5cm}}
\toprule
Concept & Neuron idx & Text generated by TopK LM \\
\midrule
none & --- & \textbf{Once upon a time,} there were two brothers, the one being the oldest and the other the youngest. [...] \\

\midrule

Work & 22:894 & \textbf{Once upon a time,} there was a huge power plant that was being pushed by a few people at one of the most challenging projects in the world.
The engine was being driven by a huge engine which was being driven by a huge to be made to be driven by a vehicle which was being driven by the power of a huge drive.
[...] 
\\

Number & 19:1439 & \textbf{Once upon a time,} seventeen: from the five:
Three from the five five: Five: fourteen:
se: Five: from from: from fourteen five: seven five: from three: fourteen: five: from from fourteen: eleven: from five: eight: from fourteen: from sixteen from fourteen eight: [...] \\

History & 18:44 & \textbf{Once upon a time,} people were called “primitive” by the Romans and by the Greeks. By the time they were called “primitive,” they were still pretty much just a bunch of different ways of thinking about things. [...] \\

Science & 17:218 & \textbf{Once upon a time,} the world was a very different place.
Scientists were still very much trying to understand how the world was changing.
But the world was changing too.
One day, in the middle of the day, a young man from the town of Kiev came to the scientists and explained what was happening. [...] \\

\bottomrule
\end{tabular}
\label{tab:steering_example}
\end{table}

%% file: figure/token_and_semantic_entropy_comparison_across_ckpt.tex
\begin{figure}[t]
\centering
\includegraphics[width=\linewidth]{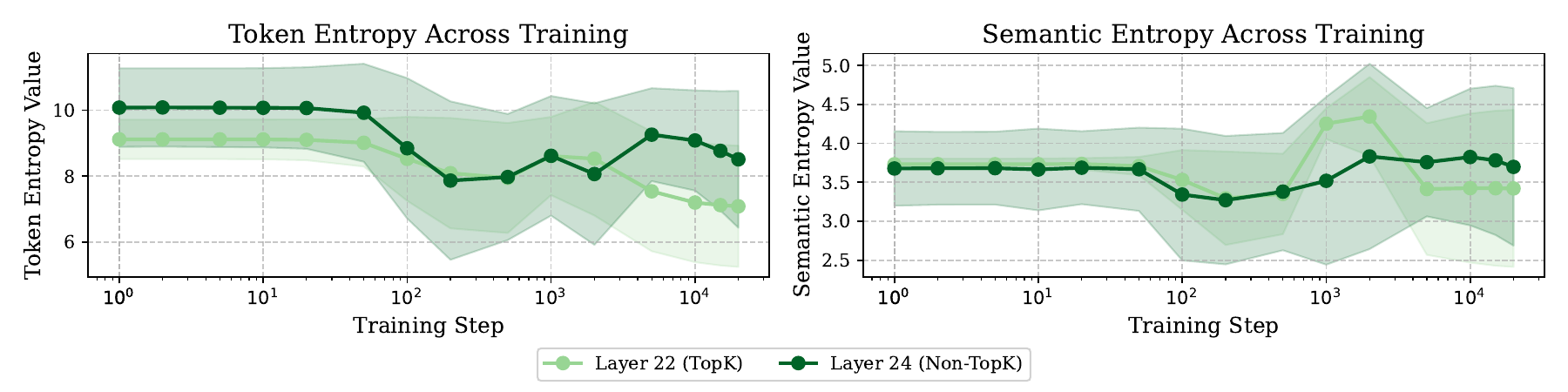}
\caption{Entropy comparison across training checkpoints for the 24-layer TopK model. 
Left: token entropy; right: semantic entropy. Curves correspond to layer 22 (last TopK layer, light green) and layer 24 (final non-TopK layer, dark green). 
The evolution of these entropy values over training reflects the progressive specialization of neuron representations.
}
\label{fig:token_and_semantic_entropy_comparison_across_ckpt}
\end{figure}

%% file: figure/layer24_neuron_activation_map_for_token_work.tex
\begin{figure}[t]
\centering
\includegraphics[width=\linewidth]{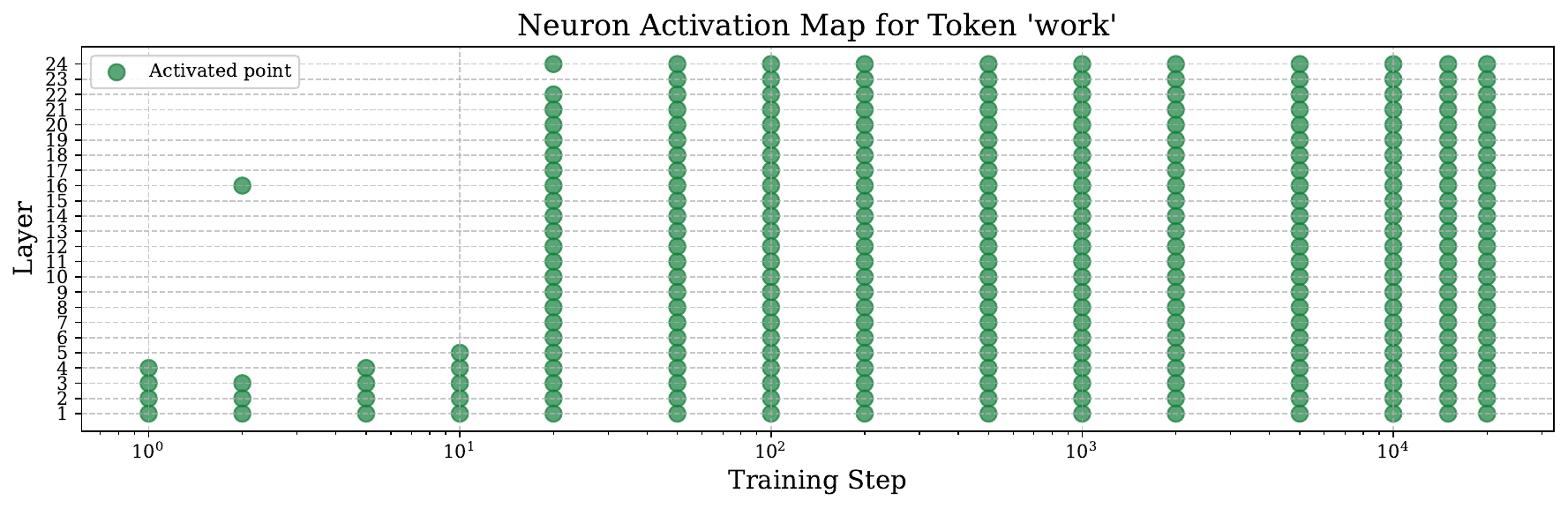}
\caption{
Visualization of the formation of a concept-specialized neuron across training checkpoints and layers.
The horizontal axis denotes training steps, and the vertical axis denotes model layer index. 
Each green marker indicates that, at the given checkpoint and layer, the neuron in hidden dimension 894 responds selectively to the token ``work.''
This spatiotemporal activation pattern reveals the developmental dynamics of neuron specialization.
}
\label{fig:layer24_neuron_activation_map_for_token_work}
\end{figure}

%% file: table/ablation_nontopk.tex
\begin{table}[t]
\centering
\small
\caption{
Ablation of the number of non‐TopK (fully dense) layers. We vary n-nontopk in an 8‐layer transformer with 1024‐dimensional hidden states and report validation, LAMBADA, and Wikitext perplexities, as well as zero‐shot accuracies on LAMBADA, ARC‐Easy, ARC‐Challenge, Winogrande, OpenBookQA, and HellaSwag.
}
\begin{adjustbox}{max width=\linewidth}
    \begin{tabular}{c|ccc|cccccc}
        \toprule
        \multirow{2}{*}{n\_nontopk} & \multicolumn{3}{c|}{Perplexity \textdownarrow} & \multicolumn{6}{c}{Accuracy \textuparrow} \\ 
        \cmidrule(lr){2-4} \cmidrule(lr){4-10} 
         & Valid. & LAMBADA & Wikitext & LAMBADA & ARC-e & ARC-c & Winogrande & OBQA & HellaSwag \\ 
        \midrule
        \multicolumn{10}{l}{\hspace{-0.22cm} {\ul hidden dim: 1024, num layer: 8}} \\[0.5em]
        - & 14.67 & 112.80 & 20.96 & 23.71 & 50.00 & 20.39 & 50.67 & 18.20 & 28.77 \\
        \midrule
        0 & 19.40 & 133.90 & 44.93 & 31.67 & 29.46 & 19.20 & 50.12 & 27.40 & 25.72 \\
        1 & 17.06 & 108.80 & 52.88 & 31.67 & 29.46 & 19.20 & 50.12 & 27.40 & 25.72 \\
        2 & 16.13 & 74.60 & 71.23 & 30.16 & 30.93 & 20.05 & 50.12 & 26.80 & 26.60 \\
        \bottomrule
    \end{tabular}
\end{adjustbox}
\label{tab:ablation_nontopk}
\end{table}

%% file: table/ablation_topk_k.tex
\begin{table}[t]
\centering
\small
\caption{
Ablation of the TopK–$k$ value. We vary the number of active units $k$ in an 8-layer transformer with 1024-dimensional hidden states and report validation, LAMBADA, and Wikitext perplexities, as well as zero-shot accuracies on LAMBADA, ARC-Easy, ARC-Challenge, Winogrande, OpenBookQA, and HellaSwag.
}
\begin{adjustbox}{max width=\linewidth}
    \begin{tabular}{c|ccc|cccccc}
        \toprule
        \multirow{2}{*}{k} & \multicolumn{3}{c|}{Perplexity \textdownarrow} & \multicolumn{6}{c}{Accuracy \textuparrow} \\ 
        \cmidrule(lr){2-4} \cmidrule(lr){5-10} 
         & Valid. & LAMBADA & Wikitext & LAMBADA & ARC-e & ARC-c & Winogrande & OBQA & HellaSwag \\ 
        \midrule
        \multicolumn{10}{l}{\hspace{-0.22cm} {\ul hidden dim: 1024, num layer: 8, topk-k:64}} \\[0.5em]
        - & 14.67 & 112.80 & 20.96 & 23.71 & 50.00 & 20.39 & 50.67 & 18.20 & 28.77 \\
        \midrule
        \phantom{00}8 & 18.75 & 116.78 & 89.42 & 30.16 & 30.93 & 20.05 & 50.12 & 26.80 & 26.60 \\
        \phantom{0}16 & 17.32 & 121.63 & 122.70 & 30.02 & 32.37 & 19.54 & 48.54 & 23.20 & 26.50 \\
        \phantom{0}32 & 16.55 & 90.30 & 76.83 & 34.66 & 31.02 & 21.59 & 50.20 & 23.80 & 26.25 \\
        \phantom{0}64 & 16.13 & 74.60 & 71.23 & 30.16 & 30.93 & 20.05 & 50.12 & 26.80 & 26.60 \\
        128 & 15.91 & 70.31 & 60.02 & 23.71 & 50.00 & 20.39 & 50.67 & 18.20 & 28.77 \\
        256 & 15.89 & 55.23 & 45.40 & 28.62 & 28.75 & 19.97 & 51.07 & 27.20 & 26.16 \\
        512 & 15.81 & 52.53 & 24.19 & 31.24 & 31.10 & 21.33 & 50.75 & 24.20 & 26.74 \\
        \bottomrule
    \end{tabular}
\end{adjustbox}
\label{tab:ablation_topk_k}
\end{table}

%% file: table/ablation_hidden_dim.tex
\begin{table}[t]
\centering
\small
\caption{
Ablation of hidden‐state dimensionality. We vary the hidden size of an 8‐layer model. We then report validation, LAMBADA, and Wikitext perplexities, as well as zero‐shot accuracies on LAMBADA, ARC‐Easy, ARC‐Challenge, Winogrande, OpenBookQA, and HellaSwag.
}

\begin{adjustbox}{max width=\linewidth}
    \begin{tabular}{c|ccc|cccccc}
        \toprule
        \multirow{2}{*}{dim} & \multicolumn{3}{c|}{Perplexity \textdownarrow} & \multicolumn{6}{c}{Accuracy \textuparrow} \\ 
        \cmidrule(lr){2-4} \cmidrule(lr){5-10} 
         & Valid. & LAMBADA & Wikitext & LAMBADA & ARC-e & ARC-c & Winogrande & OBQA & HellaSwag \\ 
        \midrule
        \multicolumn{10}{l}{\hspace{-0.22cm} {\ul num layer: 8, top-k: none}} \\[0.5em]
        1024 & 14.67 & 112.80 & 20.96 & 23.71 & 50.00 & 20.39 & 50.67 & 18.20 & 28.77 \\
        \midrule
        \multicolumn{10}{l}{\hspace{-0.22cm} {\ul num layer: 8, topp-k: 64, n-last-nontopk: 2}} \\[0.5em]
        1024 & 16.13 & 74.60 & 71.23 & 30.16 & 30.93 & 20.05 & 50.12 & 26.80 & 26.60 \\
        2048 & 12.32 & 46.61 & 46.94 & 27.71 & 33.71 & 22.01 & 49.72 & 26.60 & 26.76 \\
        3072 & 10.91 & 25.56 & 30.95 & 35.67 & 33.42 & 23.12 & 49.09 & 27.80 & 27.99 \\
        4096 & 10.16 & 42.09 & 28.53 & 26.35 & 39.39 & 21.59 & 46.57 & 27.40 & 28.68 \\
        8192& 10.28 & 32.97 & 11.63 & 28.97 & 39.06 & 24.15 & 47.91 & 27.20 & 30.45 \\
        \bottomrule
    \end{tabular}
\end{adjustbox}
\label{tab:ablation_hidden_dim}
\end{table}

%% file: figure/layer8_entropy_comparison.tex
\begin{figure}[t]
\centering
\includegraphics[width=\linewidth]{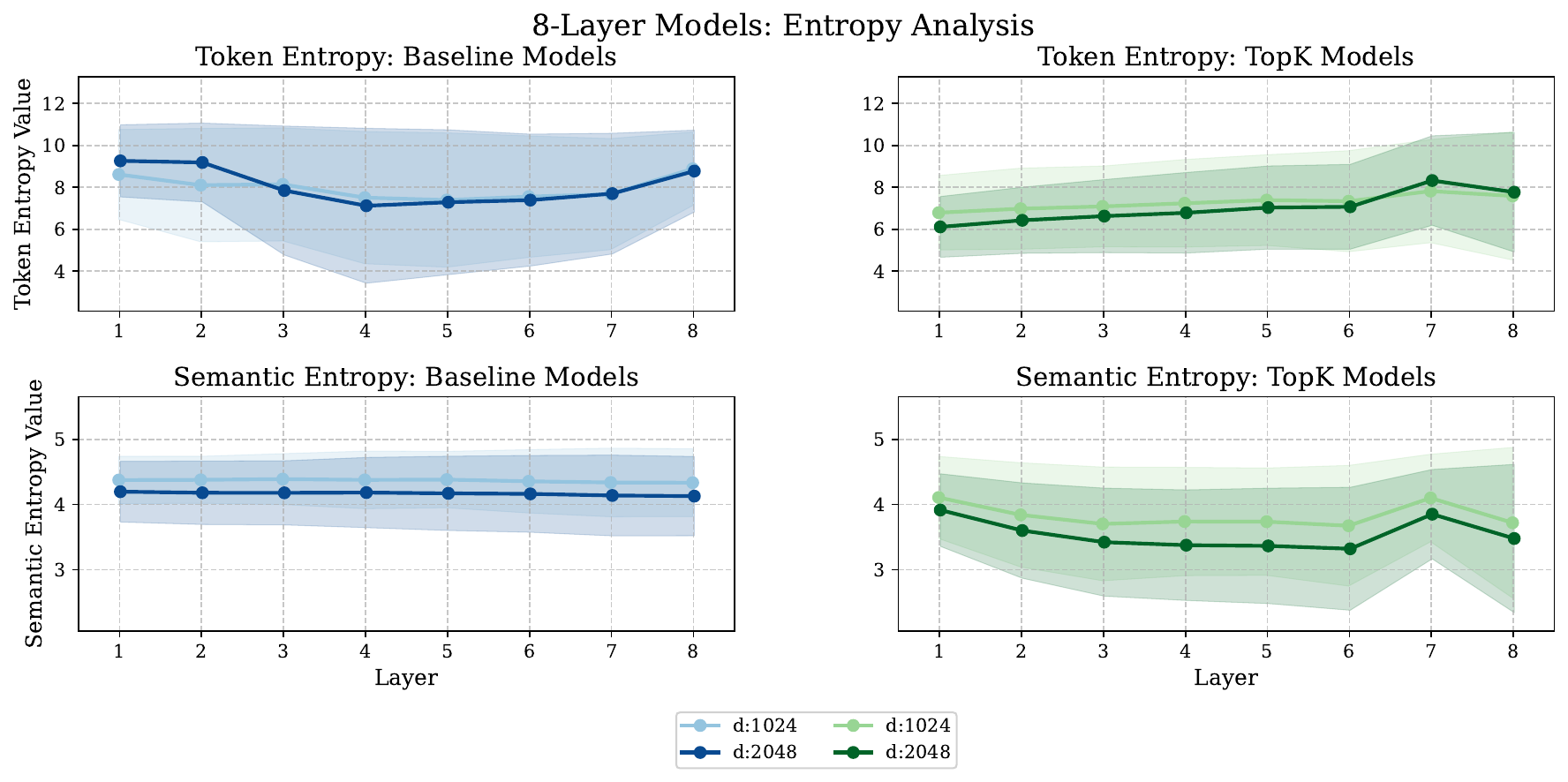}
\caption{Baseline vs TopK models (8-layer). (Left) Token Entropy. (Right) Semantic Entropy.}
\label{fig:layer8_entropy_comparison}
\end{figure}

%% file: figure/layer16_entropy_comparison.tex
\begin{figure}[t]
\centering
\includegraphics[width=\linewidth]{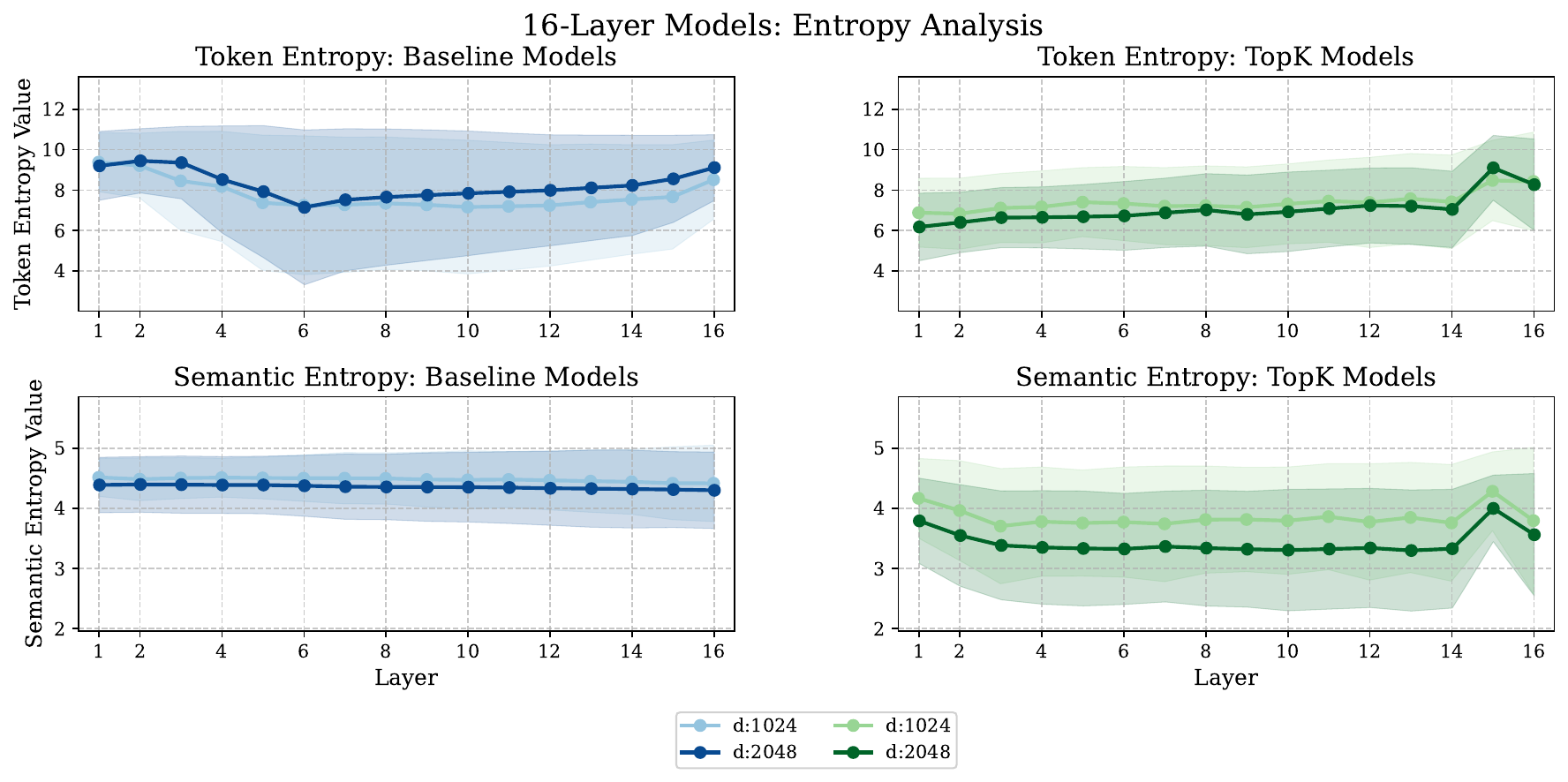}
\caption{Baseline vs TopK models (16-layer). (Left) Token Entropy. (Right) Semantic Entropy.}
\label{fig:layer16_entropy_comparison}
\end{figure}